\documentclass[twoside]{article}

\usepackage{scrwfile}
\TOCclone[\contentsname~(\appendixname)]{toc}{atoc}
\newcommand\StartAppendixEntries{}
\AfterTOCHead[toc]{%
  \renewcommand\StartAppendixEntries{\value{tocdepth}=-10000\relax}%
}
\AfterTOCHead[atoc]{%
  \edef\maintocdepth{\the\value{tocdepth}}%
  \value{tocdepth}=-10000\relax%
  \renewcommand\StartAppendixEntries{\value{tocdepth}=\maintocdepth\relax}%
}
\newcommand*\appendixwithtoc{%
  \addtocontents{toc}{\protect\StartAppendixEntries}
  \listofatoc
}

\usepackage[accepted]{aistats2025}
\usepackage{hyperref}
\usepackage{url}
\usepackage{amsmath, amsthm, amssymb}
\usepackage{algpseudocode}

\theoremstyle{plain}

\theoremstyle{definition}

\theoremstyle{remark}

\usepackage[utf8]{inputenc} 
\usepackage[T1]{fontenc}    
\usepackage{hyperref}       
\usepackage{url}            
\usepackage{booktabs}       
\usepackage{amsfonts}       
\usepackage{nicefrac}       
\usepackage{microtype}      
\usepackage{xcolor}         
\usepackage{adjustbox}
\usepackage{xcolor, color, colortbl}
\usepackage{nicematrix}
\usepackage{amsmath}
\usepackage{multicol, multirow}

\usepackage{subfigure, graphicx}
\usepackage{floatrow}

\definecolor{pink_bg}{HTML}{ffcbcb}
\definecolor{Gray}{gray}{0.95}
\definecolor{green_bg}{HTML}{f1f7b5}

\begin{document}

%

%

\twocolumn[

\aistatstitle{Type Information-Assisted Self-Supervised Knowledge Graph Denoising}

\aistatsauthor{ Jiaqi Sun$^{1,2}$ \And Yujia Zheng$^1$ \And  Xinshuai Dong$^1$ \And Haoyue Dai$^1$ \And Kun Zhang$^{1,2}$ }

\aistatsaddress{ $^1$Carnegie Mellon University \quad\quad  $^2$Mohamed bin Zayed University of Artificial Intelligence } ]

\begin{abstract}
Knowledge graphs serve as critical resources supporting intelligent systems, but they can be noisy due to imperfect automatic generation processes.
Existing approaches to noise detection often rely on external facts, logical rule constraints, or structural embeddings.
These methods are often challenged by imperfect entity alignment, flexible knowledge graph construction, and overfitting on structures. 
In this paper, we propose to exploit the consistency between entity and relation type information for noise detection, resulting a novel self-supervised knowledge graph denoising method that avoids those problems.
We formalize \textit{type inconsistency} noise as triples that deviate from the majority with respect to type-dependent reasoning along the topological structure.
Specifically, we first extract a compact representation of a given knowledge graph via an encoder that models the type dependencies of triples.
Then, the decoder reconstructs the original input knowledge graph based on the compact representation.
It is worth noting that, our proposal has the potential to address the problems of knowledge graph compression and completion, although this is not our focus.
For the specific task of noise detection, the discrepancy between the reconstruction results and the input knowledge graph provides an opportunity for denoising, which is facilitated by the type consistency embedded in our method.
Experimental validation demonstrates the effectiveness of our approach in detecting potential noise in real-world data.
\end{abstract}

\section{INTRODUCTION}
\label{sec:1_intro}
Knowledge graphs are widely used to provide expert knowledge support for intelligent systems, such as chatbots, recommendation systems, etc.~\cite{guo2020kg_rec_survey, ji2021kgsurvey, pan2024kg_llm_survey}.
A set of triples is commonly used to represent a knowledge graph, where each triple contains a head entity, a relation, and a tail entity.
Despite the fact that knowledge graphs are used as expert knowledge, they can be noisy due to the imperfect automatic generation process, which often lacks strict expert supervision~\cite{deng2023gold, ma2023learning_denoise}. 
Examples of such imperfections include computational errors during the construction of knowledge graphs from text extractions and construction bias when multiple people contribute to the cosntruction process. 
These problems can be further exacerbated when using multi-hop connections over knowledge graphs, which is particularly detrimental when precise retrieval is required~\cite{guo2020kg_rec_survey, ji2021kgsurvey, chetoui2022kg_gnn_survey}.  
In this paper, we will focus on denoising knowledge graphs by detecting potential noise in real-world data.

Before discussing related proposals, it is necessary to clarify what kind of noise we are trying to tackle in the context of knowledge graphs. 
Previous work on commonsense knowledge graphs defines noise as triples that are inconsistent with objective facts or logical reasoning based on those facts~\cite{deng2023gold}. 
However, detecting such noise is often impractical because external facts are usually missing in most knowledge graph applications, and there are entity alignment challenges if the external database is assumed to be correct~\cite{zhao2020entity_align_survey, zeng2021entity_align_survey}. 
Some rule-based approaches take into account the incompleteness, conflict, and redundancy of knowledge graphs~\cite{cheng2018rule1, belth2020normal}, but their strict definition of noise, based on a rigorous formulation of logical rules, is likely to be incompatible with the flexible construction process of real-world knowledge graphs. 
Other embedding-based methods attempt to generate a confidence score for each triple based on topological consistency, while potentially suffering from overfitting in the structures~\cite{zhang2022contrastive, zhang2023integrating_attr}. 
%
Given the above shortcomings, we try to find a more general and practical way to uncover the noisy triples, which should go beyond topological consistencies and external facts.
Consider a triple, \texttt{(head\_entity}, \texttt{relation}, \texttt{tail\_entity)}, from the NELL-995 dataset~\cite{xiong2017deeppath_nell}:
\texttt{(concept\_city\_murdoch, concept:agentcontrols, concept\_personasia\_news\_corporation)}. 
This triple does not seem sensible because Rupert Murdoch, who the triple is supposed to represent, is a businessman, which does not logically fit the entity type \texttt{city}. 
This inconsistency can be caught by noticing that the entity type \texttt{city} does not typically appear with the relation type \texttt{agentcontrols} in the data set.

Based on this discovery, in this paper, we aim to detect noise by leveraging the consistency between the type information of entities and relations. 
We define \textit{type inconsistency} noise: \textit{Given a knowledge graph, type inconsistency noise is a set of triples whose type information is inconsistent with that of the majority of existing triples or with triples derived through type-dependent reasoning.} 
This definition implies a reasonable assumption that most links are correct.
In particular, this definition of noise does not rely on any external input, allowing us to denoise the knowledge graph in a self-supervised manner.
Moreover, the type information along with the structural relations gives rise to the constraints that the legit triples should obey.
These considerations form the general idea of our proposal: a type-information assisted self-supervised knowledge graph denoising approach.

In the methodology, we introduce an auto-encoder architecture to implement this idea. 
Specifically, we first employ an encoder that extracts a compact representation of the entire knowledge graph, considering only the type information of triples and their structural dependencies. 
The decoder then reconstructs the original input knowledge graph based on this compact representation. 
The difference between the reconstructed result and the input knowledge graph provides an opportunity for denoising, which is facilitated by the type consistency embedded in our approach. It should be noted that the proposed method has the potential to address the problems of knowledge graph compression and completion, although this is not our focus~\cite{sachan2020kgcompression_survey, zamini2022kgcompletion_survey}.

In summary, our contributions are as follows: \textbf{i)} We investigate type information as a direct and effective source of revealing noisy triples in knowledge graphs. \textbf{ii)} Accordingly, we design a type information-assisted self-supervised denoising approach. \textbf{iii)} We manage to reveal noisy triples directly from real data.

\section{NOTATIONS}
\label{sec:2_pre}
\textbf{Knowledge graph and type information.}\quad
A knowledge graph consists of entities and the relations between them, typically represented as a set of triples: $K = \{(h_i,r_i,t_i)\}_{i=1}^n$, where each triple $(h_i,r_i,t_i)$ comprises a head entity $h_i$, a relation $r_i$, and a tail entity $t_i$ and there are $n$ triples in the graph. 
Each entity is from an entity token domain $V$, i.e., $\forall i\in z(n), \ h_i, t_i \in V$.
We use $z(n) = \{1,2,\cdots,n\}$ to denote the index set.
Similarly, each relation $r_i$ belongs to a relation domain $R$. 
A knowledge graph can alternatively be represented as a three-dimensional cube, i.e., $\textbf{A} \in \{0, 1\}^{|V|\times|R|\times|V|}$, where $\textbf{A}[{h,r,t}] = 1$ if $(h,r,t)\in K$, otherwise $\textbf{A}[{h,r,t}] = 0$.
Both $(h,r,t)$ and $(h_i,r_i,t_i)$ can represent a knowledge graph triple, and the latter is preferable when the index is needed.
Moreover, in this work we focus especially on the knowledge graph that comes with type information of entities, represented by a type function $c: V\vee R\mapsto C$. 
The cardinality of the entity type domain is strictly less than that of the entity domain, i.e., $|c[V]| < |V|$.
However, in most knowledge graphs, 
the type domain and the token domain are the same for relations. 
Therefore, for relations, the type function is an identity mapping.
In this paper, a knowledge graph with entity type information is denoted as $G=(K,c)$ or $G=(\textbf{A},c)$.

\textbf{Other notation convention.}\quad
We use uppercase letters to denote a set, and its member is marked as lowercase or indexed, i.e., $x$ or $x_i$. 
The cardinality of the set $X$ is denoted by $|X|$.
Bold capital letters denote matrices or a three-dimensional cube (i.e., a collection of matrices). Bold lowercase letters denote a vector. 
The entries of a given matrix are marked by using square brackets, i.e., $\textbf{X}[h,r,t]$.  
Functions are denoted using lowercase letters.
The output of a function for a singleton input is denoted as $f(x)$, and 
$f[X]$ donates the case where the input and the output are set.

\section{METHODOLOGY}
\label{sec:3_method}
\subsection{Observations and Motivation}
\begin{figure}[t]
    \centering
    \begin{adjustbox}{width=0.5\textwidth}
  \includegraphics[width=0.8\textwidth, trim={0cm 0cm 0cm 0.1cm}, clip]{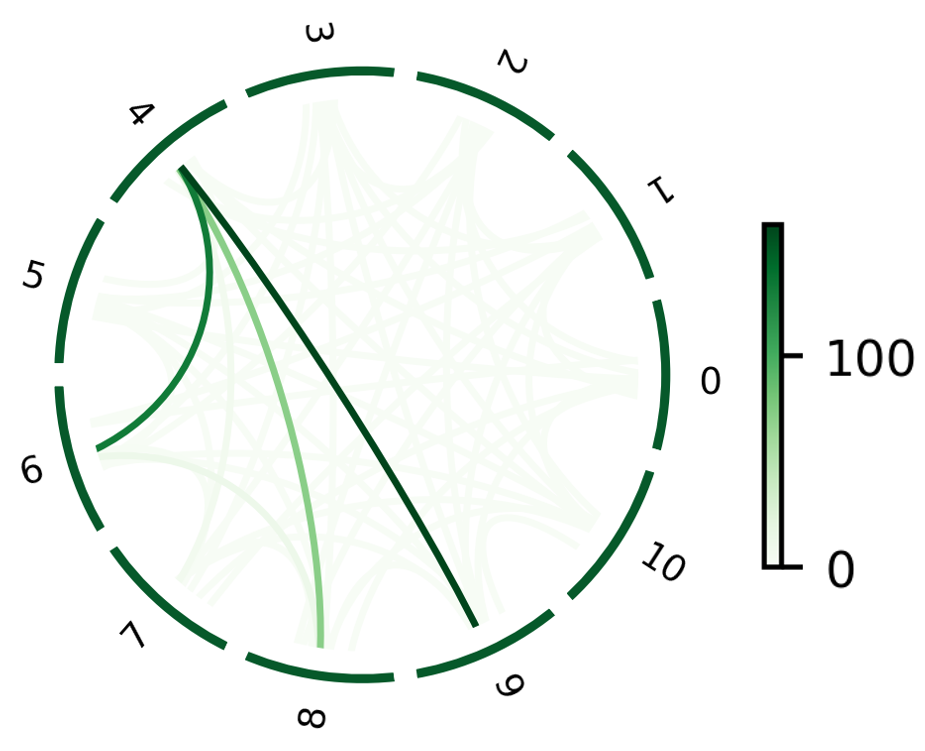}
    \caption{\small Entity types distribution w.r.t. a given relation in WN18RR.}\label{fig:3_pre_rel_dist}
    \end{adjustbox}
\end{figure}

\looseness=-1
\textbf{Type information.}\quad
First, let us elaborate on our motivation by demonstrating the correlation between type information and triples in a knowledge graph. Here, we provide quantitative evidence to illustrate the relationship between type information and the frequency of observing the corresponding triples.
Table~\ref{tab:1_pre_stats} summarizes the statistics of three widely used datasets (with detailed descriptions provided in Section~\ref{sec:4_exp}).
The cardinality of the type domain is significantly smaller than that of the token domain, as introduced in Section~\ref{sec:2_pre}.
More importantly, we calculate the proportion of legitimate triple types~(\% LTT)—defined as the existing combinations of head entity type, relation~(type), and tail entity type—relative to all possible combinations of entity and relation types in the entire dataset, as follows:
\begin{align}
    \text{\% LTT} = \frac{|\{(c(h), c(r), c(t)); (h,r,t)\in K\}|}{|c[V]|\times|c[R]|\times|c[V]|},
\end{align}
where $c$ is the type function within a given knowledge graph $G=(K,c)$.
As shown in the table, legitimate triple types represent only a small fraction of all possible triple type combinations. 
To better illustrate the concentration of type distribution, Figure~\ref{fig:3_pre_rel_dist} shows the distribution of entity types for a randomly chosen relation type in the least concentrated dataset, WN18RR.
In the figure, the darker green indicates a higher frequency of the relation type occurring between the given connected entity types. 
The results show that the possible entity types associated with a given relation type are very limited. 
Such observations are general across different datasets, which can be found in the Appendix.

This observation suggests that as the order of the link increases, the type constraints become stricter. 
For instance, given two entities with known types, the number of possible relation type combinations in the intermediate multi-hop link becomes more centrally distributed. 
Consequently, if we reasonably assume that the majority of triples in a knowledge graph are correct, and therefore their triple types are also correct, we can use this concentrated distributed type information to detect noise. 
Specifically, we can identify noisy triples when their type information deviates from the majority, as indicated by multi-hop constraints.
%
Given the feasibility of using type information as a guide for noise detection, we propose a method to model type dependencies over triples to unveil hidden noise in knowledge graphs.

\begin{table}[]
    \centering
    \begin{adjustbox}{width=\textwidth}
    \begin{tabular}{lccc}
    \toprule
     \bf{Dataset} & \bf{NELL-995} & \bf{WN18RR} & \bf{FB15k-237} \\
    \midrule
    \# entities & 75,492 & 40,943 & 14,541 \\
    \# entity types & 267 & 11 & 237 \\
    \# relations~(=types) & 200 & 11 & 237 \\
    \rowcolor{Gray}
    \% LTT & 0.13 & 21.11 & 0.05 \\
    \bottomrule
    \end{tabular}
    \caption{Type information of three real-world datasets.}\label{tab:1_pre_stats}
    \end{adjustbox}
\end{table}



\textbf{Compact triple set and denoising.}\quad
Since the topology of the knowledge graph allows for information dependencies between triples, to model the type dependencies of triples, we need to consider the type information of each triple and propagate these dependencies along the structure of the graph.
Furthermore, reduction along these topological dependencies confirms the existence of a subset of triples on which the type information of all other triples depends.
Given the above analysis, we propose to extract a compact representation of the knowledge graph, i.e., a subset of all existing triples. 
This compact set is to be inferred from the existing triples and used to infer other legitimate triples. 
In this way, noisy triples are not kept in the compact set, since they are not type-dependent on most of the existing triples. 
Moreover, for the same reason, these noisy triples cannot be constructed from the compact representation set.
Accordingly, we design a self-supervised architecture in which the encoder extracts the compact set based on type information and structural dependencies, and the decoder reconstructs the knowledge graph. The difference between the original knowledge graph triples and their reconstruction provides an avenue to detect type-inconsistency noise.

\subsection{General Proposal}
Given a knowledge graph $G=(K,\textbf{A}, c)$ containing $n$ triples, i.e., $K=\{(h_i, r_i, t_i)\}_{i=1}^n$, its three-dimensional cube representation $\textbf{A}\in\{0,1\}^{|V|\times|R|\times|V|}$, and the type information function of the entities, i.e., $c: V\mapsto C$ (here we focus on the case when the relation type domain is identical to its token domain), we want to find the most compact triples that can be used to reconstruct the original graph. 
To simplify the notation, we use $\mathcal{B}$ to denote the discrete three-dimensional space: $\{0,1\}^{|V|\times|R|\times|V|}$.
And we use $(h,r,t)$ to denote arbitrary triple if not otherwise specified.
We consider the following formulation.
\begin{subequations}
\begin{align}
& \min_{\substack{\textbf{B}\in\mathcal{B} \\ \text{supp}(\textbf{B}) \subseteq K} }||\textbf{B}||_0, \label{eq: opt_sparsity_goal}\\ 
&\text{s.t.}\quad \textbf{B} = \arg\min_{\substack{\textbf{B}\in \mathcal{B}\\ \text{supp}(\textbf{B}) \subseteq K\\ \theta\in\Theta}}
d(f_{\theta}(\textbf{B}, c), \textbf{A}). \label{eq: opt_recons_goal}
\end{align}
\end{subequations}
The above formulation corresponds to a hierarchical optimization problem~\cite{anandalingam1992hierarchical}. 
The first-level objective, i.e., $||\textbf{B}||_0$, represents the $l^0$ norm of $\textbf{B}$, and $\text{supp}(\textbf{B})$ denotes the support set of the matrix $\textbf{B}$, e.g., $\text{supp}(\textbf{A}) = K$~(definition can be found in Appendix~\ref{apd-def}). 
The reconstruction function that takes into account the structure of the graph and the type information of the triples is denoted as $f_{\theta}: \mathcal{B} \times c\mapsto \mathcal{B}$, with a learnable parameter $\theta$ searched from $\Theta$. 
The reconstruction results are measured by measuring the distance of two matrices, i.e., $d: \mathcal{B} \times \mathcal{B} \mapsto \mathbb{R}$.

To ensure that the compact representation $\textbf{B}$ does not degrade into an oversimplified matrix, e.g., a zero matrix, $\textbf{0}^{|V|\times|R|\times|V|}$, rendering it meaningless, and to confirm that the reconstruction process relies on the type information $c$ rather than overfitting to the structural information $\textbf{A}$, we consider $f$ to belong to a specific function class.
%
In other words, we want each entry in the function's output, i.e. $f_{\theta}(\textbf{B}, c)[h,r,t]$, to depend on the type information of $h, r, t$ and their topological neighbours. 
This wish is consistent with the idea of message-passing based knowledge graph embedding methods, and here we implement this desired function $f$ using Relational Graph Convolutional Networks~(R-GCN)~\cite{schlichtkrull2018rgcn}. 
In general, the $[h,r,t]-$entry of the output $f_{\theta}(\textbf{B}, c)$ is produced:
\begin{align}
    f_{\theta}(\textbf{B}, c)[h,r,t] = s(\textbf{Z}[h],\textbf{Z}[r], \textbf{Z}[t]), \label{eq: recon}
\end{align}
where $s$ denotes any function that maps the embeddings of the triple, i.e., $\textbf{Z}[h], \textbf{Z}[r], \textbf{Z}[t] \in \mathbb{R}^{d}$ into the real range of $[0,1]$. 
Consequently, the domain of the distance function $d$ extends to $[0,1]^{|V|\times|R|\times|V|} \times \mathcal{B}$.
The embeddings, i.e., $\textbf{Z}[h], \textbf{Z}[r], \textbf{Z}[t]$, are all aggregations of the propagated type information along the graph structure by using multi-layer R-GCN. 
And the first-layer representations of the entities are initialized as the type information of the entities, i.e., $\textbf{Z}[h]^{(0)} = c[h]$.
After the initialization, each R-GCN layer outputs an updated representation of entities, for example:
\begin{align}
\textbf{Z}[h]^{(l+1)} = &\sigma ( \sum_{r\in R} \sum_{j\in \mathcal{N}_{h}^{r}}  \frac{1}{c_{h,r}} \textbf{W}_r^{(l+1)} \textbf{Z}[j]^{(l)}  \\
& + \textbf{W}_0^{(l+1)} \textbf{Z}[h]^{(l)} ), 
\label{eq:rgcn_layer}
\end{align}
where $\sigma$ is the activation function. $\mathcal{N}_h^r$ represents the structural neighbor entities of a given entity $h$ that are connect by a specific relation type $r$.
Thus, the parameter space of $f$, i.e., $\Theta$, consists of all the transformation parameter matrices $\{\textbf{W}_{r}^{(l)}; r\in R, l=1,\cdots,L\}$, given a hyper-parameter $L$ representing the depth of the network, the self linear transformation $\textbf{W}_0^{l}$ for each layer, a learnable representation for each relation type, i.e., $\textbf{Z}[r]\in \mathbb{R}^{d}; r\in R$, and the possible parameters of $s$.

%

%
%

\subsection{Parameterization and Overall Objective}

\textbf{Parameterized search.}\quad
As presented in Eq.~(\ref{eq: opt_sparsity_goal}) and (\ref{eq: opt_recons_goal}), the search space for $\textbf{B}$ is discrete and vast, that is, $(\textbf{B}\in\mathcal{B}) \wedge (\text{supp}(\textbf{B}) \subseteq K)$, making direct optimization of this space challenging. 
To facilitate computation, we consider parameterizing the search space, transforming the originally discrete space into a continuous one.
To achieve this goal, recall that $\textbf{B}$ is the most compact representation of $\textbf{A}$.
Hence, we naturally assume that the entry value of $\textbf{B}[h,r,t]$, which indicates whether the corresponding triple $(h,r,t)$ belongs to the compact representation of the knowledge graph, should be inferred from the neighboring connections and the type information $c$. 
In simple terms, a triple is retained in the compact representation of a knowledge graph if it semantically aligns with the local connections.
We implement this by implementing a masking function on the given knowledge graph, i.e. $m: \mathcal{B} \times c\mapsto \mathcal{B}$, which determines whether a triple should be retained in the compact representation. 
We use a design similar to the reconstruction function, as shown below:
\begin{align}
    m_{\phi}(\textbf{A}, c)[h,r,t] = s(\textbf{H}[h],\textbf{H}[r], \textbf{H}[t]),  \label{eq: mask}
\end{align}
where $\phi$ is the learnable parameter in the masking function, and its parameter space includes all parameters involved in calculating the embeddings denoted by $\textbf{H}$ and the possibly potential parameters in $s$.
The vector $\textbf{H}[h]\in \mathbb{R}^d$ is recursively calculated using Eq.~(\ref{eq:rgcn_layer}), initialized in the same way as above.
In the implementation of $s$, we deploy a Multi-Layer Perceptron~(MLP) to enhance the flexibility of inferring the compact set from all triplets.
Also note that the parameters of $f_{\theta}$ and $m_{\phi}$ are independent, as indicated by the different symbols used, i.e., $\textbf{Z}$ and $\textbf{H}$ are different, although their internal function classes are identical.

Given this efficient parameterization of the search space, the original optimization problem, as demonstrated in Eqs.(\ref{eq: opt_sparsity_goal}) and (\ref{eq: opt_recons_goal}), is modified into the following form:
\begin{subequations}
\begin{align}
& \min_{\phi \in \Phi} || m_{\phi}(\textbf{A}, c) ||_0, \label{eq: opt_sparsity_goal_mod}\\ 
& \text{s.t.}\quad m_{\phi}(\textbf{A}, c) = \arg\min_{\substack{\phi \in \Phi \\ \theta\in\Theta}} 
d\left(f_{\theta}\left(m_{\phi}(\textbf{A}, c\right), c\right), \textbf{A}). \label{eq: opt_recons_goal_mod}
\end{align}
\end{subequations}
\textbf{Asymptotic discretization.}\quad
The parameterization outlined above facilitates continuous optimization using gradient descent. 
However, it is worth noting that in the modified framework, i.e., Eqs.(\ref{eq: opt_sparsity_goal_mod}) and (\ref{eq: opt_recons_goal_mod}), both the reconstruction function and the masking function produce continuous output, creating a discrepancy between the original problem and its modified version. 
To address this, we employ suitable techniques to discretize the output.

First, we apply a sigmoid transformation to the output of both functions to polarize the values.
Furthermore, we utilize Gumbel-Softmax~\cite{jang2017gumbel} to further discretize the output of the masking function. 
This is particularly important because the output of the masking function serves as the input to the reconstruction function, which requires stricter discretization.
For instance, if $q_i$ represents an entry of the output of the original masking function, i.e., Eq.~(\ref{eq: mask}), the actual input to the reconstruction function is given by:
\begin{align}
    \frac{\exp(\frac{\log(\frac{1}{1+\exp(-q_i)} + p_i)}{\tau} )}
    {\exp(\frac{\log(\frac{1}{1+\exp(-q_i)} + p_i)}{\tau} ) + \exp(\frac{\log(\frac{1}{1+\exp(-(1-q_i))} + p_i')}{\tau})},
    \label{eq: gumble}
\end{align}
where $p_i$ and $p_i'$ are independently sampled from a Gumbel distribution with a location parameter of $0$ and a scale parameter of $1$, and $\tau$ is a hyper-parameter controlling the sparsity of the outputs.

\textbf{The overall optimization objective and the sparsity constraint.}\quad
We consider the following objective to solve the constrained optimization problem formulated in Eqs.\ref{eq: opt_sparsity_goal_mod} and \ref{eq: opt_recons_goal_mod}.
\begin{align}
\min_{\substack{\phi \in \Phi \\ \theta\in\Theta}} 
d\left(f_{\theta}\left(m_{\phi}(\textbf{A}, c\right), c\right), \textbf{A}) + \gamma \rho(m_{\phi}(\textbf{A}, c) ), \label{eq: final_obj}
\end{align}
where $\rho(m_{\phi}(\textbf{A}, c) )$ is a sparsity regularizer with a non-zero hyper-parameter $\gamma$ controlling the strength, for replacing the original $l^0$ norm constrain, because the original one is the discrete component, making continuous optimization of the whole problem difficult.
Considering the relevant discussions on sparsity constraints~\cite{ng2024sparse}, we adopt the minimax concave penalty regularizers~\cite{zhang2010mcp}, where both the $l^1$ and $l^2$ norms are involved:
\[
\rho(\textbf{X}[i,j,k]) = 
    \begin{cases}
    \lambda|\textbf{X}[i,j,k]|-\frac{\textbf{X}[i,j,k]^2}{2\alpha},\text{if } |\textbf{X}[i,j,k]|\leq \alpha\lambda,\\
    \frac{\alpha\lambda^2}{2}, \text{   otherwise}.\\
    \end{cases}
\]    

\subsection{Denoising Based on the Discrepancy}
Given the objective formulated by Eq.~\eqref{eq: final_obj}, the optimized functions $m_{\phi}$ and $f_{\theta}$ can be obtained.
Respectively, $f_{\theta^*}$ is the general mechanism that generates the links by the minimal fundamental set of triples, which is marked by $m_{\phi^*}(\textbf{A}, c)$.
Obviously. 
Our idea is that noisy triples should not be contained in the reconstruction $f_{\theta^*}(m_{\phi^*}(\textbf{A}, c), c)$.
Therefore, we can detect noisy triples simply by comparing the discrepancy between the original input $\textbf{A}$ and the reconstruction, with a chosen threshold $0.5$.
More specifically, given a triple $(h,r,t)$ that appears in the original knowledge graph, that is, $\textbf{A}[h,r,t] = 1$, the noise label $y((h, r, t))$ is determined in the following form:
\[
y((h,r,t)) = 
    \begin{cases}
    1,& \text{if } f_{\theta^*}\left(m_{\phi^*}(\textbf{A}, c) \right)[h,r,t]\geq 0.5  \\
    0,& \text{otherwise}.\\
    \end{cases}
\] 
Similarly, the task of completing the knowledge graph can be accomplished by introducing the triples that have entry values greater than or equal to the threshold. 
And the compression task is already done by obtaining the fundamental representation of the knowledge graph, i.e., $m_{\phi^*}(\textbf{A}, c)$.
The sensitivity analysis of choosing the threshold is provided in the Appendix.
In addition, the diagram demonstration of the training and inference phases can be found in Figure~\ref{fig:model-train} and Figure~\ref{fig:model-denoise}.

\section{DISCUSSIONS}
\textbf{The analogy to single image denoising.}\quad
To make it more intuitive, let us draw the analogy between our knowledge graph denoising framework and single image denoising techniques~\cite{quan2020self2self, huang2021neighbor2neighbor, ko2023self2self+}. 
In these methods, a self-supervised approach is employed, where no reference image is provided, and thereby, the original image is used as input to the denoiser, which outputs an image of the same size. 
The assumption is that local patches of the image, regardless of their location, should preserve the same noise pattern. 
This allows the denoiser to be trained on multiple local patches.
Similarly, our knowledge graph denoising proposal assumes a shared noise pattern at the type information level, which abstracts the token-level information we observe. 
In this way, ``local patches'' can be constructed using the different token-level embodiments of the abstract type information. We would like to add that the proposed framework is expected to help in knowledge graph compression and completion as well, although we do not focus on them~\cite{zhang2020kgcompression1, sachan2020kgcompression_survey, zamini2022kgcompletion_survey}.

\textbf{Sample size and the type abstraction.}\quad
Just as in image denoising, where a larger image provides more local patches to train a more stable and generalizable denoising model (while avoiding converging to identity mapping), knowledge graph denoising also benefits from a sufficiently large knowledge graph. 
Thus, our experiments focus on real-world knowledge graphs to ensure a sufficient scale for effective denoising.
It is crucial that the abstraction of the knowledge graph should not be too high for an effective denoising, as overly abstract type information can make it difficult for the model to fit the input graph information. 
Finally, we would like to note that the compact representation proposed in our approach is essential for denoising tasks, since it reduces the chance that the model performs identity mapping by storing the most informative part of the graph, ensuring that noise is not preserved.

\section{EXPERIMENTS}
\label{sec:4_exp}
To evaluate our proposal for detecting noise, we focus on uncovering noise directly from the original dataset, given that the actual distributions of noise in real-world knowledge graphs are typically unknown. 
This approach, which aims to reveal noise in real-world datasets, has been relatively unexplored in existing research. 
Our experiments focus on two sets of questions.
\textbf{RQ1}: Is the identified noise reasonable? Can other knowledge graph embedding methods achieve similar noise detection?
\textbf{RQ2}: Does the proposed method avoid fitting noisy data? How do the individual components of our proposed method contribute to this attribute?
\textbf{RQ3}: Is our proposal robust to mild corruption of type information?
Due to page limit, we save i) noise detection verification of general setting (i.e., uncover the additionally added abnormal triples from the original knowledge graph) and the results on larger-scale datasets, ii) hyper-parameter sensitivity analysis, iii)
computational complexity examination, and iv) preliminary evaluation of our proposal on knowledge graph completion and compression task in the Appendix.

\subsection{Settings}
\textbf{Dataset and baselines.}\quad
We conduct experiments mainly on three datasets: WN18RR~\cite{bordes2013translating_wn18rr}, FB15k-237~\cite{toutanova2015observed_fb15k}, and NELL-995~\cite{xiong2017deeppath_nell}. 
In addition, the larger-scale datasets are also considered, namely, ogbl-biokg~\cite{hu2020open}, DBpedia~\cite{lehmann2015dbpedia}, and Yago~\cite{mahdisoltani2013yago3}, which are put in the Appendix due to page limit.
The entity type information for the WN18RR and FB15k-237 datasets was obtained from the publicly available OpenKE project\footnote{\href{https://github.com/thunlp/OpenKE}{https://github.com/thunlp/OpenKE}}, where the entity type information is extracted based on the relation type. 
For FB15k-237, we obtained the entity types from the corresponding text descriptions. 

For the baselines, here we choose embedding-based denoising approaches for fair comparisons and also due to the limit of page. For other types of baseline, please refer to the Appendix.
Among them, the most related are the MPNN-based embedding approaches: R-GCN~\cite{schlichtkrull2018rgcn}, KBGAT~\cite{nathani2019kbgat}, and the denoising methods GCNN~\cite{jia2019gcnn}, and TRUST~\cite{neil2018interpretable_noise}.
We also compare MLP-enhanced distance measurement methods, namely DistMult~\cite{yang2015distmult} and ConvE~\cite{dettmers2018conve}; Their superiority compared to vanilla representation space modeling methods has been revealed~\cite{li2023message}; therefore, we do not additionally include those methods as baselines.
In these methods, negative sampling is usually employed to enhance robustness.
Because we are concerned with revealing noise in real-world datasets, rather than revealing additional noise added to the original dataset, these suggestions for combating robustness are already embodied in the knowledge graph embedding methods described above.

\textbf{Implementations.}\quad
Like most MPNN-based knowledge graph embedding approaches that do not rely on strict logical rules, we typically process triples by appending a reverse version of each. 
This practice fully leverages the asymmetric dependencies between triples. For example, the example we give in Section~\ref{sec:1_intro}, \texttt{(concept\_personasia\_news\_corporation, concept:agentcontrols\_reverse, concept\_city\_murdoch)} is also included in the training set of these approaches. 
In our implementation, we adopt this practice, considering that type dependencies are also asymmetric. This modification further strengthens the type dependencies our method relies on.
For the hyper-parameters of the masking and reconstruction functions, we use the optimized settings suggested by recent work~\cite{li2023message}, and we provide detailed settings in Appendix~\ref{apd-hyper}. 
Since our proposed method is an auto-encoder architecture based on R-GCN, the abbreviation RAE (i.e., R-GCN Auto-Encoder) denotes our proposal.
More implementation details regarding RAE and baselines are provided in the Appendix~\footnote{The link to our code repository is \href{https://github.com/sajqavril/Code-Repo-for-R-GCN-Auto-Encoder.git}{https://github.com/sajqavril/Code-Repo-for-R-GCN-Auto-Encoder.git}.}.

\textbf{Evaluation metrics.}\quad
We first focus on revealing the noise triples from real-world data, where the true error label is unknown. In this case, we report the number of uncovered noisy triples for \textbf{RQ1} and \textbf{RQ3}, which is computed as: $\#E = |\{f_{\theta^*}\left(m_{\phi^*}(\textbf{A}, c) \right)[h,r,t]\geq 0.5 : \textbf{A}[h,r,t] = 1\}|$.
In the Appendix, we elaborate the metrics we use for evaluating the performance of artificially added error triples (i.e., RQ i)), when all the observed triples are assumed to be correct and the noise labels are known, including true negative rate given a bandit.

\begin{table}[t]
    \centering
    \begin{adjustbox}{width=\textwidth}
    \begin{tabular}{lccc}
    \toprule
    \textbf{Method} & \textbf{NELL-995} & \textbf{WN18RR} & \textbf{FB15k-237} \\
    \midrule
    DistMult & $1.50 \pm 1.58$ & $0.50 \pm 0.85$ & $3.20 \pm 1.23$ \\
    ConvE    & $1.90 \pm 0.24$ & $1.22 \pm 0.24$ & $4.60 \pm 3.24$ \\
    KBGAT    & $0.50 \pm 0.76$ & $2.83 \pm 1.72$ & $15.00 \pm 22.49$ \\
    R-GCN    & $1184.50 \pm 675.65$ & $442.00 \pm 124.45$ & $343.00 \pm 87.88$  \\ 
    GCNN  & $0.65 \pm 0.27$ &  $2.25 \pm 1.90$ & $12.23 \pm 3.44$\\
    TRUST & $1.84 \pm 0.73$ &  $3.15 \pm 1.55$ & $13.51 \pm 15.43$ \\
    \cellcolor{Gray}RAE~(Ours)    & \cellcolor{Gray}$25.00 \pm 1.00$ &  \cellcolor{Gray}$49.25 \pm 42.17$ & $120.67 \pm 5.03$\cellcolor{Gray} \\ 
    \bottomrule
    \end{tabular}
    \end{adjustbox}
    \caption{\small Comparisons of the number of detected noise among different approaches~($\#E$).}
    \label{tab:exp1_noise_detection}
\end{table}

\begin{table*}[h]
\killfloatstyle
    \centering
    \caption{\small Detected noise from the NELL-995 dataset~(best viewed in color).}
    \begin{adjustbox}{width=0.8\columnwidth}
    \begin{tabular}{ccc}
\toprule
\textbf{Head Entity}  & \textbf{Relation} & \textbf{Tail Entity}  \\
\midrule
person\_larry\_page & agentcontrol & university\_google \\ 
stateorprovince\_idaho & stateorprovinceisborderedbystateorprovince & stateorprovince\_south\_dakota \\
\rowcolor{pink_bg} 
personasia\_news\_corporation & agentcontrol & city\_murdoch \\ 
\rowcolor{Gray}
\multicolumn{3}{c}{Eight non-noisy triples have been omitted due to page limitations.}\\  
\rowcolor{pink_bg} 
ceo\_robert\_iger & agentcontrol & person\_disney \\ 
company\_apple002 & headquarteredin & city\_london\_city  \\ 
\rowcolor{pink_bg} 
personmexico\_m\_s & agentcontrol & director\_stuart\_rose  \\ 
\rowcolor{Gray}
\multicolumn{3}{c}{Four non-noisy triples have been omitted due to page limitations.}\\
\rowcolor{pink_bg} 
profession\_parts & proxyfor & book\_new \\ 
\rowcolor{pink_bg}
sportsleague\_irl & agentcontrol & personmexico\_tony\_george  \\ 
\rowcolor{Gray}
\multicolumn{3}{c}{Four non-noisy triples have been omitted due to page limitations.}\\
\rowcolor{pink_bg} 
movie\_repo & synonymfor & airport\_epel  \\ 
recordlabel\_roc\_a\_fella & organizationterminatedperson & ceo\_damon\_dash  \\
\bottomrule
\end{tabular}
    \end{adjustbox}
    \label{tab:exp4-rgcnae-sp0.1-ep14}
\end{table*}

\subsection{Experimental Results on Real-World Data}
\looseness=-1
\textbf{RQ1: noise detection and case study on NELL-995.}\quad 
We employ the reconstruction discrepancy-based noise detection method outlined in Section~\ref{sec:3_method} to identify potential noise in three real-world datasets. 
Table~\ref{tab:exp1_noise_detection} shows that our proposed method consistently detects a stable number of noisy triples with lower variance, except for WN18RR. 
A possible reason is that this dataset is characterized by overly abstract type information (as shown in Table~\ref{tab:1_pre_stats}), which poses challenges to our proposal to effectively fit the triples.
In contrast, the baselines either vary widely in the number of detected noisy triples, such as R-GCN, or converge on nearly all the triples, leaving little room for denoising, such as DistMult, KBGAT, GCNN, and TRUST. 
We also include the DistMult training process in Appendix~\ref{apd-exp1-process} to confirm its fit to all triples.
Table~\ref{tab:exp4-rgcnae-sp0.1-ep14} shows the noise detected from NELL-995 by our proposal. 
For readability, we remove the ``concept'' prefix.
Obvious noise is highlighted (some non-noise triples are omitted; please refer to Appendix~\ref{apd-full-tab} for the complete list). 
%
In particular, most noise triples are detected by their reverse counterparts. 
For example, the noisy triple \texttt{(ceo\_robert\_iger}, \texttt{agentcontrol}, \texttt{person\_disney)} is originally captured by its reverse counterpart \texttt{(person\_disney}, \texttt{agentcontrol\_reverse}, \texttt{ceo\_robert\_iger)}. 
This finding further supports our approach of reversing links to better utilize the type constraints contained in the triples.

\begin{table}[t]
    \centering
    \begin{adjustbox}{width=\textwidth}
    \begin{tabular}{cccc}
    \toprule
        Corrupt. ratio & NELL-995 & WN18RR & FB15k-237 \\
         \midrule
        w/o & $25.00 \pm 1.00$ &  $49.25 \pm 42.17$ & $120.67 \pm 5.03$ \\ 
        w/ 0.01\% & $23.37 \pm 1.37$ & $45.73 \pm 51.32$ & $118.94 \pm 3.11$\\
        w/ 0.1\%  & $26.09 \pm 2.02$ & $44.25 \pm 52.87$ & $124.85 \pm 6.36$\\
        w/ 1.0\% & $21.57 \pm  4.29$ & $52.36 \pm 58.66$ & $110.61 \pm 10.30$\\
         \bottomrule
    \end{tabular}
    \caption{\small Robustness w.r.t. corrupted type labels~($\# E$).}
    \label{tab:rebuttal-robustness-type}
    \end{adjustbox}
\end{table}

\begin{figure*}
    \centering
    \includegraphics[width=0.95\textwidth, trim={1cm 0cm 0cm 0cm}, clip]{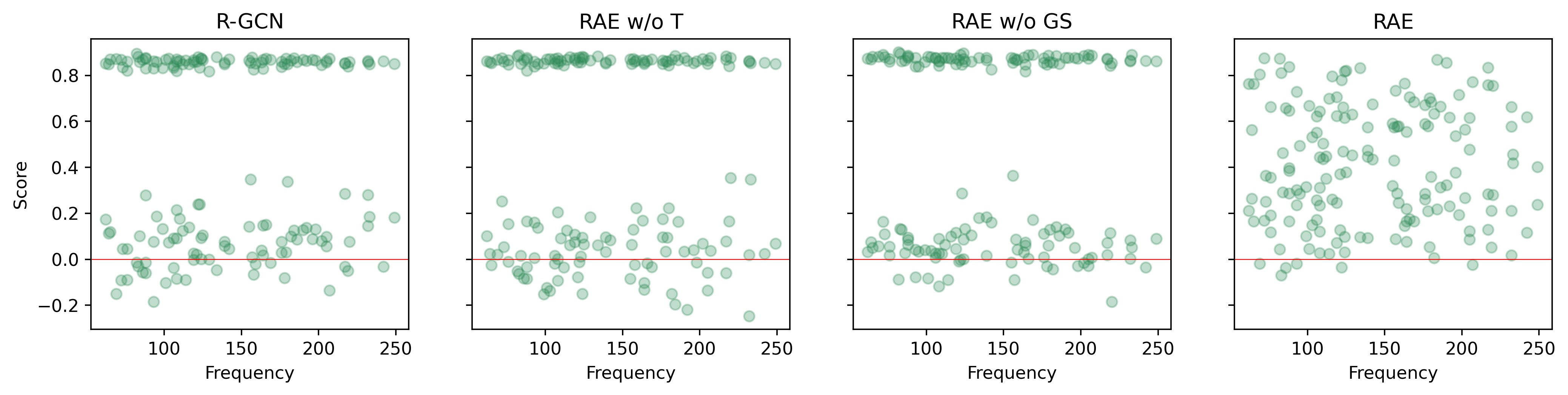}
    \vspace{-0.3cm}
    \caption{\small Comparisons of different methods on fitting triples with various type frequencies ~(\#F).}
    \label{fig: exp2_fit_noise}
\end{figure*}

\looseness=-1
\textbf{RQ2: avoiding fitting on noise.}\quad
As we analyze in Section~\ref{sec:3_method}, the triples are distributed centralized with respect to the types. Generally speaking, if a triple has a type that agrees with the majority of the triples' type, then it is less likely to be a noisy triple. 
Given that the ground truth of noisy triples is unknown in real-world datasets, we indirectly demonstrate how the model avoids fitting noise by showing how well it fits the more frequently occurring types of triple.
In Figure~\ref{fig: exp2_fit_noise}, we compare the performance of R-GCN, our proposed RAE, RAE without type information guidance (RAE w/o T), and RAE without Gumbel-Softmax (RAE w/o GS).
RAE w/o T treats each entity as a different token like R-GCN, maintaining the auto-encoder framework. RAE w/o GS removes the Gumbel-Softmax component.
Each point in the figure represents a triple type, and the y-axis score for each type is the average score of all triples of that type minus the scores of 10 randomly sampled negative triples (created by replacing the tail entities).
For example, given a triple type $(c_h,r,c_t)$, the score is calculated as:
\begin{align}
    \#F = \frac{1}{|S|} \sum_{(h,r,t)\in S}(\hat{\textbf{A}}[h,r,t] - \frac{1}{10}\sum_{(h',r',t')\in S'}\hat{\textbf{A}}[h',r',t']), 
\end{align}
where $S=\{(h,r,t) : c[h] = c_h, c[r]=r, c[t]=c_t\}$ marks the set of all the triples that have the type of $(c_h,r,c_t)$, and $\hat{\textbf{A}}=f_{\theta^*}(m_{\phi^*}(\textbf{A}, c), c)$ is the reconstruction.
$(h',r',t')$ is the corresponding corrupted negative triple, each of which is from $S'$, whose entity or relation is randomly corrupted.
From the results, we find that only the complete RAE model effectively captures high-frequency type triples.
The comparison between R-GCN and RAE w/o GS shows that the asymptotic discretization achieved through the Gumbel-Softmax contributes to better capturing the type constraints among the triples.
The procedure of drawing this picture is put in the Appendix.

\textbf{RQ3: robustness for corrupted type information}
Since we explicitly reveal the type-inconsistent noise in our method, the robustness of our proposal with respect to type corruption is worth investigating.
Consequently, in Table~\ref{tab:rebuttal-robustness-type} we compare the performance ($\#E$) when different fractions of the entitytype labels are corrupted, from $0$ to $1.0\%$, where we find that the proposal is quite robust to the acceptable fraction of corruption.

\section{RELATED WORK}
\label{sec:7_related}
\textbf{Knowledge graph denosing.}\quad
There are mainly three types of methods to deal with noise in knowledge graphs.
The first is the use of logical rules, which considers the noisy triples as those that violate the predefined rules~\cite{galarraga2013rule2, pellissier2017rule3, cheng2018rule1}. 
However, these methods make too strict assumptions about the correct triples of following the logical reasoning rules, which may not be compatible with real-world knowledge graph constructions. 
Except for such hard rule constraints, some work exploits soft rules, that is, inductive combinations of the rules defined by the type or attributes information of the triples~\cite{belth2020normal}. 
In both cases, it is still possible that there are unknown additional constraints between edges other than explicit rules, so it is desirable to develop an automatic approach to discover all such constraints from the data.
%
Recently, a framework for denoising common-sense knowledge graphs based on pre-train language models~\cite{deng2023gold} has emerged, aiming to use external knowledge to denoise common-sense knowledge graphs. 
For such methods to work, except for the assumption that the additional supervision input is correct, it requires a high quality of entity alignment for the external knowledge to help, and the cost of obtaining such knowledge is not trivial. 
The third class of methods uses knowledge graph embedding techniques to give rise to a confidence score (sometimes called trustworthiness) of the triples, which is generated based on the structure of the knowledge graph~\cite{paulheim2015emb1,neil2018interpretable_noise,jia2019emb2,xu2022emb3,zhang2022contrastive, zhang2023integrating_attr}. These works either rely on topological information without semantic support, which may lead to overfitting on the structure if a given knowledge graph (e.g., see the results shown in Table~\ref{tab:exp1_noise_detection}). Or, expensive attributes of the entities are required~\cite{zhang2023integrating_attr}, which is not feasible for most real-world scenarios.

\textbf{Remark.}\quad
From the related work discussed, except for those that require external knowledge, we can see that many of them are trying to reveal noise by imposing regularities among the triples, by logical rules~\cite{cheng2018rule1, belth2020normal}, or by extracting the invariant among the changes~\cite{dong2023active, zhang2022contrastive}.
In addition, the importance of semantic information beyond the topological relationships captured by embedding-based approaches~\cite{zhao2019embedding_noise} has been pointed out~\cite{zhang2023integrating_attr, belth2020normal}. 
Therefore, our proposal integrates these merits and uses type information to reveal the noisy triples by enforcing type-consistent constraints on knowledge graphs.
Finally, our explicit assumptions about the noise we are trying to tackle allow us to evaluate directly on real knowledge graphs without error labels, which is harder than revealing random false positive triples.

\section{CONCLUSIONS}
\label{sec:5_con}
This work investigates the feasibility of incorporating type information to identify noise in knowledge graphs. 
Given the definition of type-inconsistency noise, we propose a self-supervised approach based on type information to denoise knowledge graphs. 
This approach leverages the readily available type information in datasets and demonstrates its ability to avoid fitting noisy triples and effectively detect potential noise in real-world datasets.

\textbf{Limitations and future work.}\quad
Our proposal still faces challenges in distinguishing isolated facts, such as triples that are independent of existing triples, and noise of type inconsistency, which is consistent with our analysis. More research is needed to differentiate these facts, possibly by analyzing the distributional differences between the two sets.
In scenarios where the embedded type information in the dataset is limited—either in quantity or quality—we plan to explore multi-hop combinations of relations and entity types to enhance the support from type information. In addition, the interpretability of the masking and reconstruction functions needs to be improved. 

\section{ACKNOWLEDGEMENTS}
All the authors would like to thank the anonymous reviewers for their generous feedback. We would also like to acknowledge the support from NSF Award No. 2229881, AI Institute for Societal Decision Making (AI-SDM), the National Institutes of Health (NIH) under Contract R01HL159805, and grants from Quris AI, Florin Court Capital, and MBZUAI-WIS Joint Program.

\bibliographystyle{apalike}
\bibliography{cr-main}




\section*{Checklist}



 \begin{enumerate}

 \item For all models and algorithms presented, check if you include:
 \begin{enumerate}
   \item A clear description of the mathematical setting, assumptions, algorithm, and/or model. [Yes] 
   \item An analysis of the properties and complexity (time, space, sample size) of any algorithm. [Yes]
   \item (Optional) Anonymized source code, with specification of all dependencies, including external libraries. [Yes]
 \end{enumerate}

 \item For any theoretical claim, check if you include:
 \begin{enumerate}
   \item Statements of the full set of assumptions of all theoretical results. [Yes] 
   \item Complete proofs of all theoretical results. [Yes]
   \item Clear explanations of any assumptions. [Yes]    
 \end{enumerate}

 \item For all figures and tables that present empirical results, check if you include:
 \begin{enumerate}
   \item The code, data, and instructions needed to reproduce the main experimental results (either in the supplemental material or as a URL). [Yes]
   \item All the training details (e.g., data splits, hyperparameters, how they were chosen). [Yes]
         \item A clear definition of the specific measure or statistics and error bars (e.g., with respect to the random seed after running experiments multiple times). [Yes]
         \item A description of the computing infrastructure used. (e.g., type of GPUs, internal cluster, or cloud provider). [Yes]
 \end{enumerate}

 \item If you are using existing assets (e.g., code, data, models) or curating/releasing new assets, check if you include:
 \begin{enumerate}
   \item Citations of the creator If your work uses existing assets. [Yes]
   \item The license information of the assets, if applicable. [Yes]
   \item New assets either in the supplemental material or as a URL, if applicable. [Not Applicable]
   \item Information about consent from data providers/curators. [Yes]
   \item Discussion of sensible content if applicable, e.g., personally identifiable information or offensive content. [Not Applicable]
 \end{enumerate}

 \item If you used crowdsourcing or conducted research with human subjects, check if you include:
 \begin{enumerate}
   \item The full text of instructions given to participants and screenshots. [Not Applicable]
   \item Descriptions of potential participant risks, with links to Institutional Review Board (IRB) approvals if applicable. [Not Applicable]
   \item The estimated hourly wage paid to participants and the total amount spent on participant compensation. [Not Applicable]
 \end{enumerate}

 \end{enumerate}


\onecolumn
\appendix
\label{apd}
\appendixwithtoc
\setcounter{theorem}{0}

\section{MORE EXPERIMENTS ON DENOISING TASK}

\subsection{Experimental Settings}





\subsubsection{Hyper-Parameters Settings}
\label{apd-hyper}

In the experiments, we refer to the framework proposed by the analytical work between MPNN-based and MLP-based knowledge graph embedding framework~\cite{li2023message}. 
The hyper-parameters of the models are listed as below:

\begin{itemize}
\item R-GCN:
\begin{itemize}
    \item FB15k-237: negative\_sampling=10, score\_function=DistMult, num\_blocks=100, learning\_rate=0.001, batch=512, l2=0, num\_workers=3, gcn\_layer=2, hidden\_dropout=0.1.
    \item NELL-995: negative\_sampling=10, score\_function=DistMult, num\_blocks=100, learning\_rate=0.001, batch=512, l2=0, num\_workers=3, gcn\_layer=2, hidden\_dropout=0.1.
    \item WN18RR: negative\_sampling=10, score\_function=DistMult, num\_blocks=100, learning\_rate=0.001, batch=512, l2=0, num\_workers=3, gcn\_layer=2, hidden\_dropout=0.1.
\end{itemize}
\item RAE~(Ours):
\begin{itemize}
    \item FB15k-237: negative\_sampling=10, score\_function=DistMult, num\_blocks=100, learning\_rate=0.001, batch=512, l2=0, num\_workers=3, gcn\_layer=2, hidden\_dropout=0.1, sparsity\_constrain=0.5.
    \item NELL-995:negative\_sampling=10, score\_function=DistMult, num\_blocks=100, learning\_rate=0.001, batch=512, l2=0, num\_workers=3, gcn\_layer=2, hidden\_dropout=0.1, sparsity\_constrain=0.5.
    \item WN18RR: negative\_sampling=10, score\_function=DistMult, num\_blocks=100, learning\_rate=0.001, batch=512, l2=0, num\_workers=3, gcn\_layer=2, hidden\_dropout=0.1, sparsity\_constrain=0.5.
\end{itemize}
\item KBGAT:
\begin{itemize}
    \item FB15k-237: negative\_sampling=0, score\_function=ConvE, learning\_rate=0.001, batch=512, l2=0, num\_workers=3, gcn\_layer=2, num\_heads=2, hidden\_dropout=0.3.
    \item NELL-995: negative\_sampling=0, score\_function=ConvE, learning\_rate=0.001, batch=512, l2=0, num\_workers=3, gcn\_layer=2, num\_heads=8, hidden\_dropout=0.3.
    \item WN18RR: negative\_sampling=0, score\_function=ConvE, learning\_rate=0.001, batch=512, l2=0, num\_workers=3, gcn\_layer=2, num\_heads=8, hidden\_dropout=0.3.
\end{itemize}
\item MLP-DistMult:
\begin{itemize}
    \item FB15k-237: negative\_sampling=0, score\_function=DistMult, learning\_rate=0.0001, batch=512, l2=0, num\_workers=3, layer=2, hidden\_dropout=0.05.
    \item NELL-995: negative\_sampling=0, score\_function=DistMult, learning\_rate=0.0001, batch=512, l2=0, num\_workers=3, layer=2, hidden\_dropout=0.05.
    \item WN18RR: negative\_sampling=0, score\_function=DistMult, learning\_rate=0.0001, batch=512, l2=0, num\_workers=3, layer=2, hidden\_dropout=0.05.
\end{itemize}
\item MLP-ConvE:
\begin{itemize}
    \item FB15k-237: negative\_sampling=0, score\_function=ConvE, learning\_rate=0.001, batch=512, l2=0, num\_workers=3, layer=1, hidden\_dropout=0.3.
    \item NELL-995: negative\_sampling=0, score\_function=ConvE, learning\_rate=0.001, batch=512, l2=0, num\_workers=3, layer=1, hidden\_dropout=0.3.
    \item WN18RR: negative\_sampling=0, score\_function=ConvE, learning\_rate=0.001, batch=512, l2=0, num\_workers=3, layer=1, hidden\_dropout=0.3.
\end{itemize}
\end{itemize}

In addition, for our method, the hyper-parameters $\alpha$ and $\lambda$ in the minimax concave regularizer are set to $10$ and $1$, respectively. 
For the strength of the sparsity constrain, we conducted comparisons between $\{0.1, 0.5, 1.0\}$ without  observing significant differences, and we chose $0.5$ as the default setting when no specific note is provided.

In addition, for all the reported numbers, we ran the experiments five times independently with random seeds chosen from $\{41504, 42, 0, 1, 2\}$ and calculated the mean and variance values. 
Exceptions were made for the case study, as it is challenging to calculate the average and variance across different runs. 
However, we additionally provide the results from other runs, see the Tables below, where we found the differences between different runs were not significant, as demonstrated in Table~\ref{tab:exp1_noise_detection}.

\subsubsection{Computation Architecture and Optimization Methods}
On the hardware side, all experiments were conducted on a machine equipped with a 24 vCPU Intel(R) Xeon(R) Platinum 8352V CPU @ 2.10GHz and 2 GPUs (RTX 4090, each with 24GB of memory).
On the software side, CUDA 12.1 was used, and the computational platform primarily consisted of PyTorch 2.1.0 and Python 3.10 running on Ubuntu 22.04.
All optimization problems were solved using the Adam optimizer with a default weight decay of 0.00005. 
The learning rate varied across different baselines and datasets, and detailed learning rate settings are provided in the appended hyper-parameters settings section~\ref{apd-hyper}.

\subsection{Comprehensive Visualization of the Type Distribution}
\label{apd-more-type}
\begin{figure}[h]
    \centering
    \includegraphics[width=\textwidth, trim={7cm 3cm 6cm 4cm}, clip]{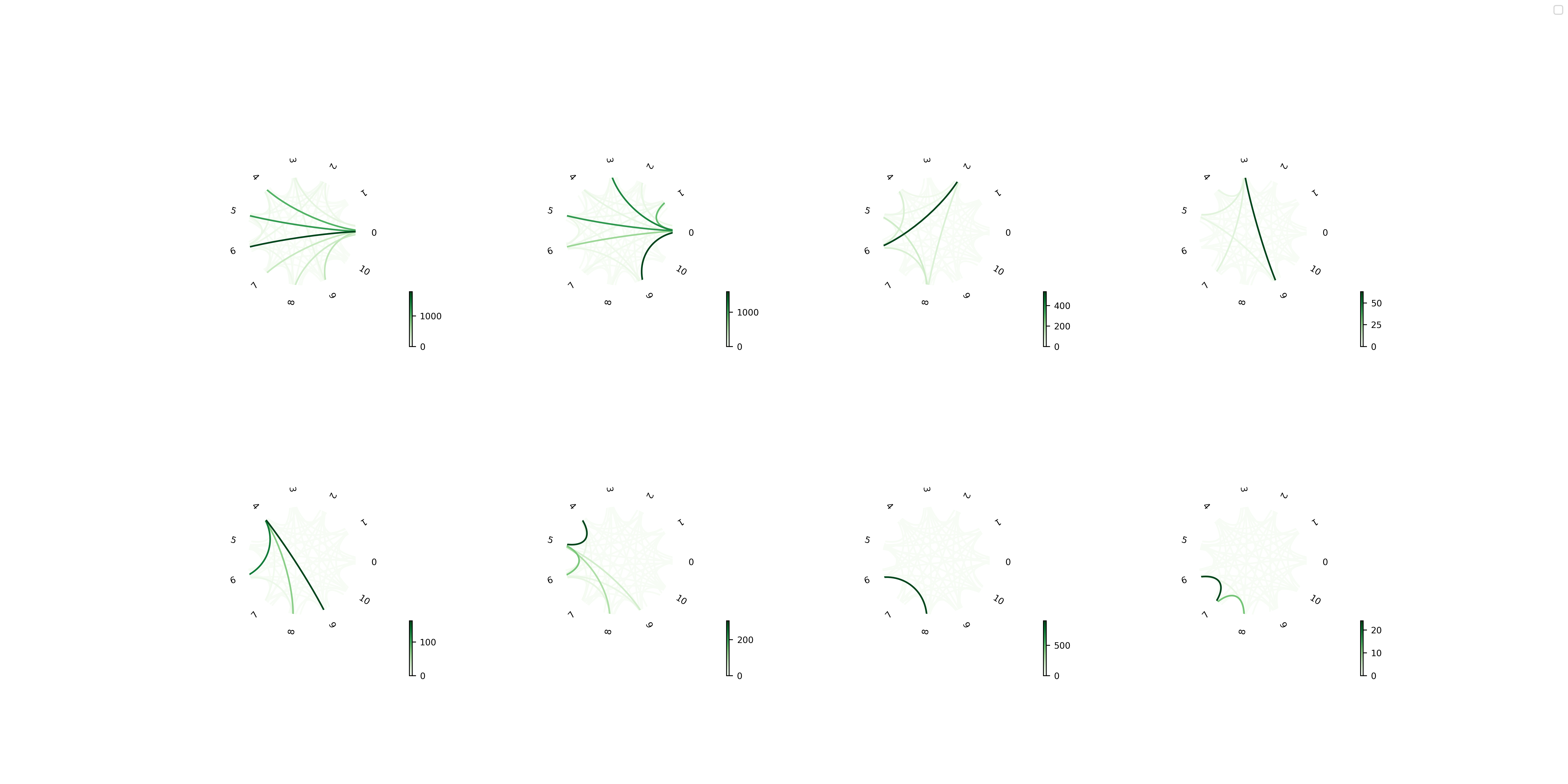}
    \label{fig:pre_rel_dist_more_type}
    \caption{Eight relation type distributions on WN18RR dataset.}
\end{figure}

Except for Figure~\ref{fig:3_pre_rel_dist}, which we showed to motivate our proposal in the main content, Figure~\ref{fig:pre_rel_dist_more_type} shows more relation type distributions corresponding to the entity types, where we can see the legitimate relation types are tightly distributed across all possible combinations of entity types.

\subsection{Converging Process of Compared Methods}
\label{apd-exp1-process}

\begin{figure}[h]
    \centering
    \subfigure[Convergence process of DistMult]{
    \includegraphics[width=\textwidth]{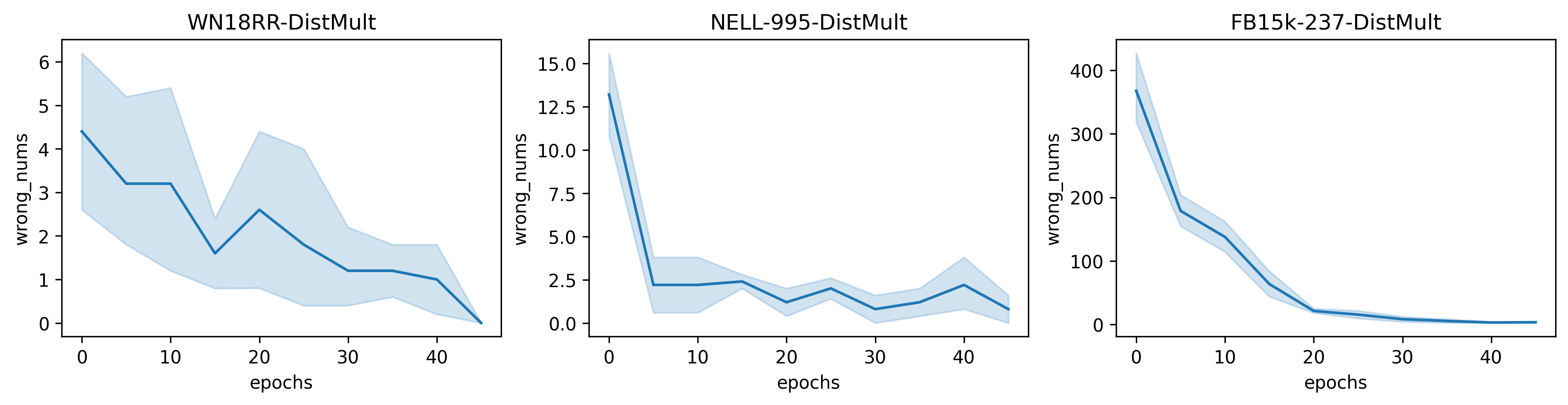}
    \label{fig:apd-distmult-converge}
    }

    \centering
    \subfigure[Convergence process of KBGAT]{
    \includegraphics[width=\textwidth]{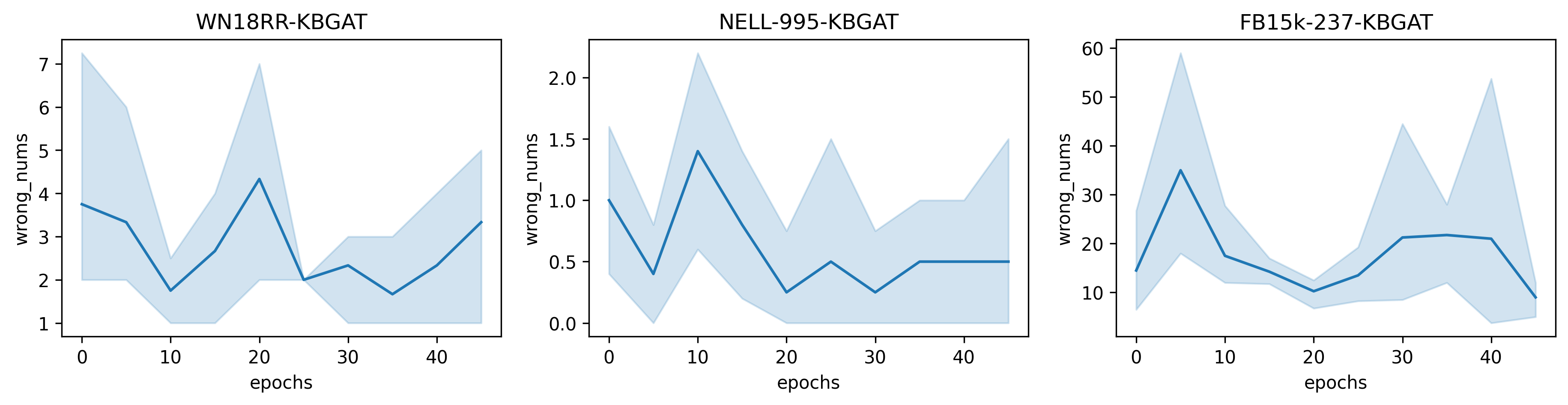}
    \label{fig:apd-kgbat-converge}
    }

    \centering
    \subfigure[Convergence process of RAE]{
    \includegraphics[width=0.7\textwidth]{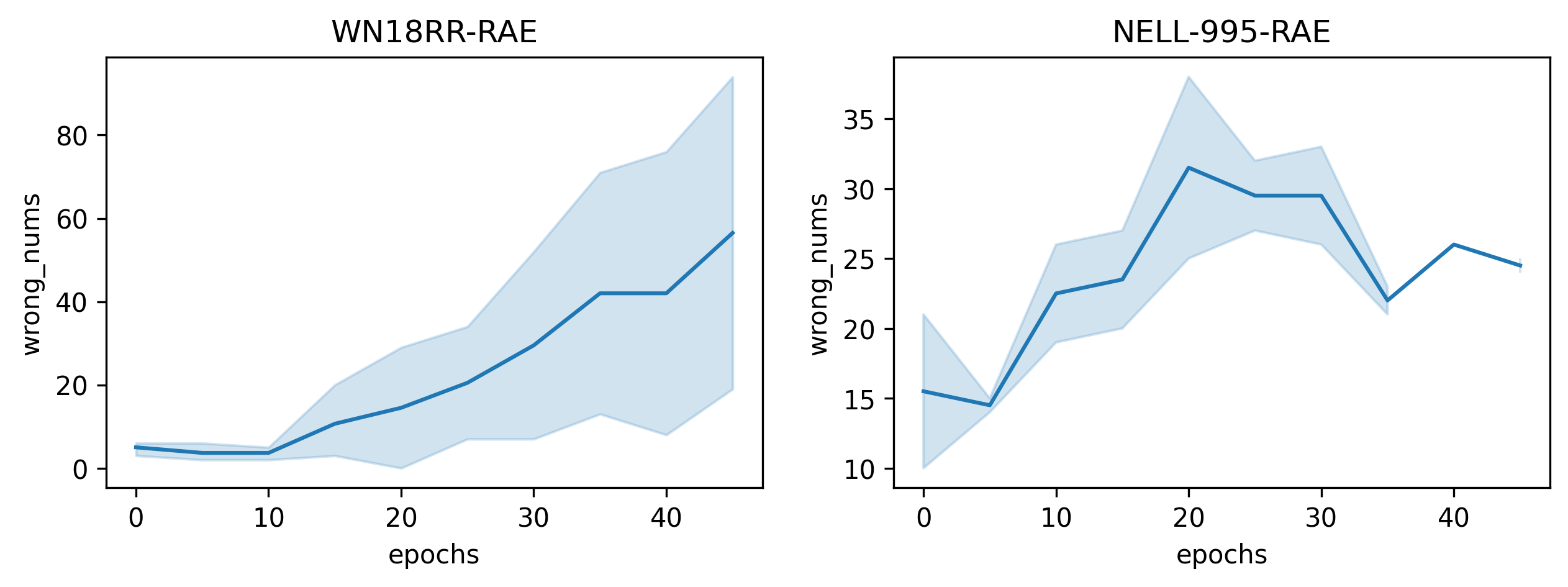}
    \label{fig:apd-rgcnae-converge}
    }
    \caption{Comparisons of convergence processes of different models}
    \label{fig:apd-converge-three}
\end{figure}

As shown in Figure~\ref{fig:apd-distmult-converge}, Figure~\ref{fig:apd-kgbat-converge} and Figure~\ref{fig:apd-rgcnae-converge}, the convergence processes of the difference methods are compared when applied to the three data sets, i.e. WN18RR, NELL-995 and FB15k-237. From the results we can observe that both DistMult and KBGAT suffer from an overfitting problem, as the number of unfitted triples tends to zero, especially DistMult. This confirms our previous claim that embedding-based methods tend to fit all topological information and cannot effectively use higher-level semantic information, e.g. type information, to detect noise.

\subsection{Full Set of Detected Noisy Triples from NELL-995}
\label{apd-full-tab}
\begin{table}[h]
\killfloatstyle
    \centering
    \caption{Full set of detected noise from NELL-995 dataset~(better view in color).}
    \begin{adjustbox}{width=\columnwidth}
    \begin{tabular}{ccc}
\toprule
\textbf{Head Entity}  & \textbf{Relation} & \textbf{Tail Entity} \\
\midrule
person\_larry\_page & agentcontrol & university\_google  \\ 
stateorprovince\_idaho & stateorprovinceisborderedbystateorprovince & stateorprovince\_south\_dakota \\
\rowcolor{pink_bg} 
personasia\_news\_corporation & agentcontrol & city\_murdoch  \\ biotechcompany\_delta\_air\_lines\_inc & organizationterminatedperson & journalist\_richard\_anderson \\ 
magazine\_gucci & organizationterminatedperson & ceo\_domenico\_de\_sole\\ 
stateorprovince\_idaho & stateorprovinceisborderedbystateorprovince & visualizablescene\_washington  \\ 
stateorprovince\_idaho & stateorprovinceisborderedbystateorprovince & stateorprovince\_oregon \\ 
sportsteam\_detroit\_tigers & teamplayssport & sport\_baseball  \\ 
person\_wendy001 & persongraduatedfromuniversit & university\_state\_university \\
musicartist\_the\_jam & agentcollaborateswithagent & male\_bruce\_foxton  \\ 
bank\_bear\_stearns & organizationterminatedperson & ceo\_james\_cayne\\
\rowcolor{pink_bg} 
ceo\_robert\_iger & agentcontrol & person\_disney \\ 
company\_apple002 & headquarteredin & city\_london\_city \\ 
\rowcolor{pink_bg} 
personmexico\_m\_s & agentcontrol & director\_stuart\_rose \\ 
school\_shichahai\_school & agentcontrol & island\_zhang  \\
website\_technorati & competeswith & blog\_google \\
sportsteam\_arizona\_state\_sun\_devils & agentcollaborateswithagent & personmexico\_ncaa  \\ 
city\_erie & proxyfor & stateorprovince\_pennsylvania  \\
\rowcolor{pink_bg} 
profession\_parts & proxyfor & book\_new  \\ 
\rowcolor{pink_bg}
sportsleague\_irl & agentcontrol & personmexico\_tony\_george \\ 
professor\_richard\_stallman & personleadsorganization & nonprofitorganization\_free\_software\_foundation  \\ 
website\_cnn\_\_fox & competeswith & newspaper\_times  \\ 
ceo\_william\_r\_\_klesse & agentcontrol & petroleumrefiningcompany\_valero\_energy  \\ company\_pimco & agentcontrol & wine\_gross  \\ 
website\_youtube & competeswith & website\_yahoo  \\
\rowcolor{pink_bg} 
movie\_repo & synonymfor & airport\_epel  \\ 
recordlabel\_roc\_a\_fella & organizationterminatedperson & ceo\_damon\_dash  \\
\bottomrule
\end{tabular}
    \end{adjustbox}
    \label{tab:full}
\end{table}

Table~\ref{tab:full} shows all the detected noisy triples.


\subsection{Parameter Sensitivity Analysis and Robustness Study}
We additionally study the sensitivity of our proposal, with respect to the depth of the R-GCN, and the strength of the sparsity constraint. 
From Table~\ref{tab:rebuttal-sensitivity-depth} we can see that the different values (e.g. the choice of 0.1, 0.5 and 1.0) of $\gamma$ do not affect the noise detection performance of the RAE in different data sets. Table~\ref{tab:rebuttal-sensitivity-sparsity} shows how the depth of the R-GCN model used in both the encoder and the decoder can affect the noise detection performance of the RAE. From the results we can see that as long as the expressiveness of the model becomes sufficient, e.g. two layers for the datasets used in the experiments, the noise detection remains quite stable thereafter.

\begin{table}[h]
    \centering
    \begin{tabular}{cccc}
    \toprule
        $L$ & NELL-995 & WN18RR & FB15k-237 \\
         \midrule
        1 & $44.92 \pm 16.06$ &  $113.48 \pm 32.90$ & $90.65 \pm 15.62$ \\ 
        2 & $25.00 \pm 1.00$ &  $49.25 \pm 42.17$ & $120.67 \pm 5.03$ \\ 
        3 & $22.70 \pm 1.43$ &  $58.21 \pm 31.69$ & $114.55 \pm 4.58$ \\ 
        4 & $23.80 \pm 0.61$ &  $41.77 \pm 29.09$ & $126.54 \pm 3.90$ \\ 
         \bottomrule
    \end{tabular}
    \caption{Sensitivity study on the depth of R-GCN $L$}
    \label{tab:rebuttal-sensitivity-depth}
\end{table}
\begin{table}[h]
    \centering
    \begin{tabular}{cccc}
    \toprule
        $\gamma$ & NELL-995 & WN18RR & FB15k-237 \\
         \midrule
        0.1 & $23.32 \pm 1.57$ &  $43.77 \pm 22.91$ & $131.44 \pm 4.81$ \\ 
        0.5 & $25.00 \pm 1.00$ &  $49.25 \pm 42.17$ & $120.67 \pm 5.03$ \\ 
        1.0 & $21.56 \pm 2.03$ &  $58.74 \pm 36.22$ & $122.62 \pm 7.40$ \\ 
         \bottomrule
    \end{tabular}
    \caption{Sensitivity study on the sparsity constraint $\gamma$}
    \label{tab:rebuttal-sensitivity-sparsity}
\end{table}

\section{DIAGRAM DEMONSTRATION OF THE PROPOSED RAE}

The training and inference phases are demonstrated in Figure~\ref{fig:model-train} and Figure~\ref{fig:model-denoise}. Due to some adjustments in the structure of the appendix, the Figure~7 and Figure~8 that we put in the main content are actually Figure~4(a) and Figure~(b) here.

\begin{figure}[h]
\centering
\subfigure[Training Phase of RAE]{
\label{fig:model-train}
\includegraphics[width=\textwidth, trim={0cm 3cm 0cm 0cm}, clip]{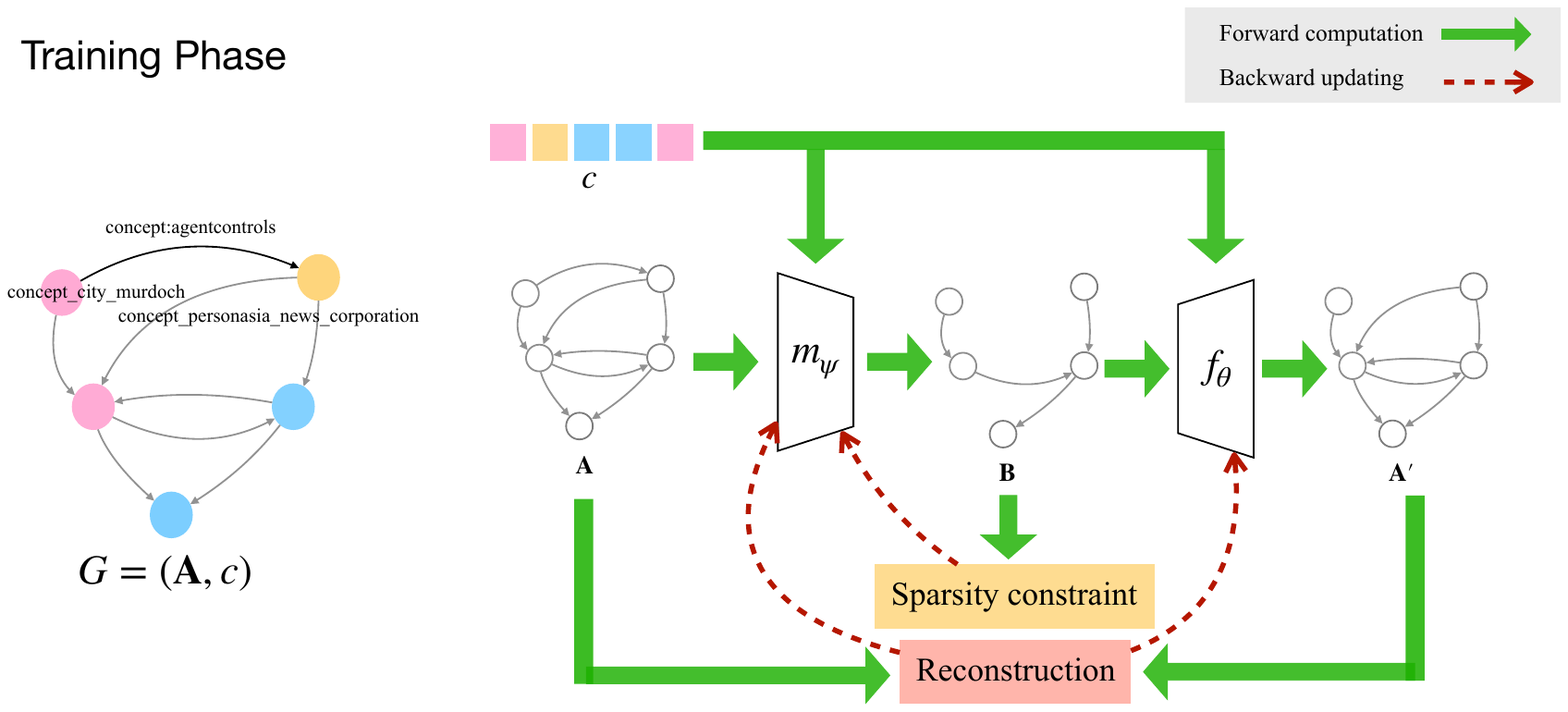}
}
\subfigure[Denoising Phase of RAE]{
\label{fig:model-denoise}
\includegraphics[width=\textwidth, trim={0cm 3cm 0cm 0cm}, clip]{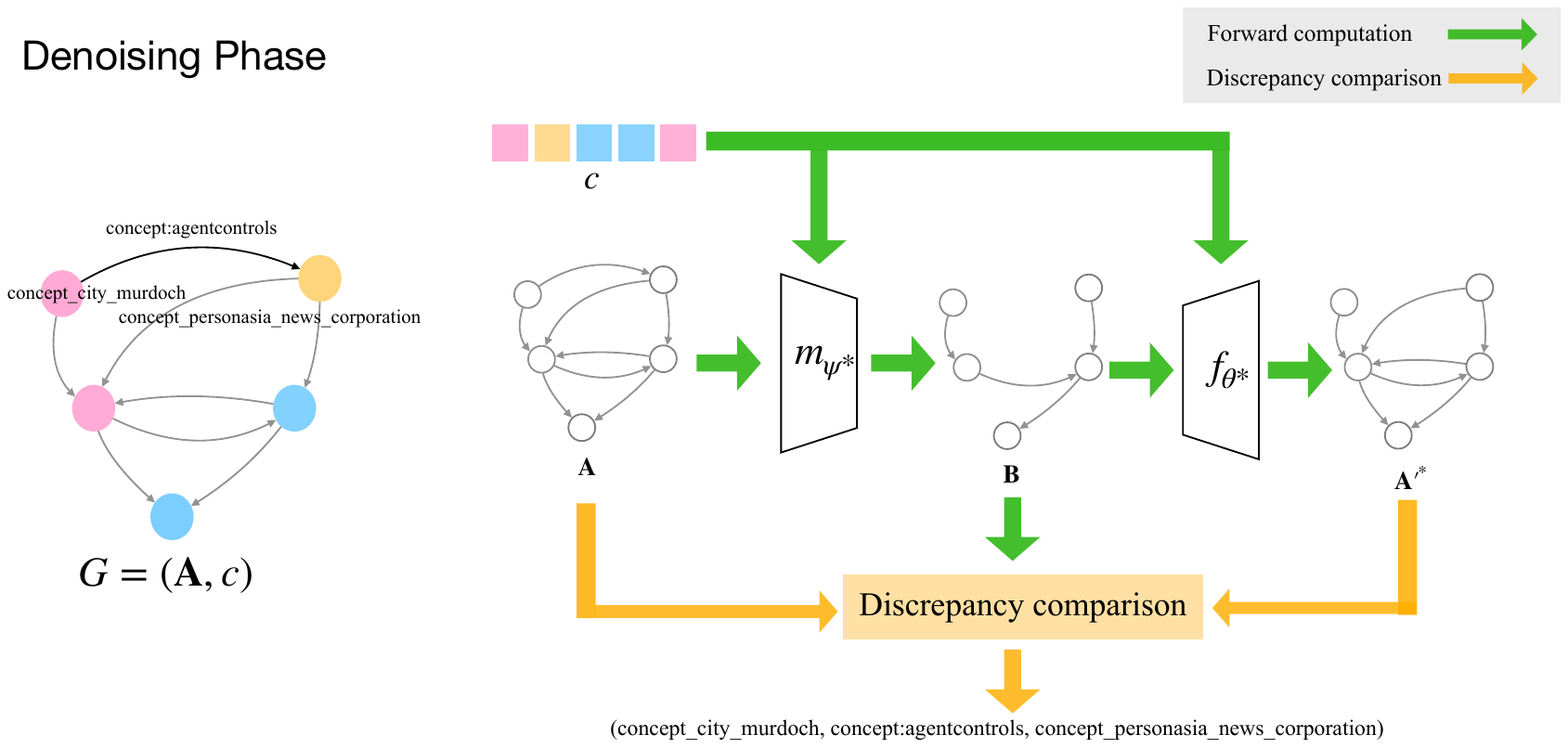}
}
\caption{Diagram demonstration of the training and inference phases of our proposal RAE.}
\label{fig:model-training-and-denoise}
\end{figure}


\section{BOARDER IMPACTS}
This work has the potential to enhance the reliability of AI applications that incorporate knowledge graphs as external expert knowledge support. However, since the interpretability of the model is not confirmed, it might be used to deliberately reject some facts by a third party. Therefore, the usage of this method should ensure the integrity of the original dataset, preventing any malicious manipulations.

\end{document}